# scientific reports

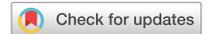

# OPEN  Applications of nature-inspired metaheuristic algorithms for tackling optimization problems across disciplines

Elvis Han Cui[1]✉, Zizhao Zhang[1,2], Culsome Junwen Chen[3] & Weng Kee Wong[1,4]✉

Nature-inspired metaheuristic algorithms are important components of artificial intelligence, and are increasingly used across disciplines to tackle various types of challenging optimization problems. This paper demonstrates the usefulness of such algorithms for solving a variety of challenging optimization problems in statistics using a nature-inspired metaheuristic algorithm called competitive swarm optimizer with mutated agents (CSO-MA). This algorithm was proposed by one of the authors and its superior performance relative to many of its competitors had been demonstrated in earlier work and again in this paper. The main goal of this paper is to show a typical nature-inspired metaheuristic algorithmi, like CSO-MA, is efficient for tackling many different types of optimization problems in statistics. Our applications are new and include finding maximum likelihood estimates of parameters in a single cell generalized trend model to study pseudotime in bioinformatics, estimating parameters in the commonly used Rasch model in education research, finding M-estimates for a Cox regression in a Markov renewal model, performing matrix completion tasks to impute missing data for a two compartment model, and selecting variables optimally in an ecology problem in China. To further demonstrate the flexibility of metaheuristics, we also find an optimal design for a car refueling experiment in the auto industry using a logistic model with multiple interacting factors. In addition, we show that metaheuristics can sometimes outperform optimization algorithms commonly used in statistics.

A reason behind the very successful and ubiquitous applications of AI and machine learning is the rapid development of clever and more effective metahheuristic algorithms for optimization purposes[1–5]. One such class is the class of nature-inspired metaheuristic algorithms that include genetic algorithm (GA), differential evolution (DE) and particle swarm optimization (PSO), among many others. Each of these algorithms has been widely tested for optimizing different types of complex objective functions successfully across disciplines. The more popular and exemplary ones have many variants, which are modified versions or improvements of the original version. For example, the variant may converge faster, make the original algorithm less prone to premature convergence, or has greater chance of extricating itself from a local optimum. Ease of availability of codes in R, Matlab and Python to run metaheuristics greatly facilitate its use and popularity in practice. For example, the website https://pyswarms.readthedocs.io/en/latest/ houses a comprehensive set of PSO tools written in Python[6]. More recently,[7] provided a high-level Python package for selecting machine learning algorithms and their parameters using PSO. Hybridized algorithms that creatively combine suitable algorithms, metaheuristic or not, can also markedly increase the performance of a metaheuristic algorithm; see details and applications in[8–10].

There are many applications of nature-inspired metaheuristic algorithm across disciplines. For example, PSO, being an exemplary nature-inspired algorithm, is widely used to tackle problems due to COVID-19[11–14]. There are many monographs on nature-inspired metaheuristic algorithms at various levels, see, for example,[15–18]. Some are targeted to specific disciplines; for example, applications in building energy power and storage systems[4], agriculture[19], chemical engineering[20], or for feature selection[21] with numerous applications in finance and reinforce learning, to name a few. Overview papers on metaheuristics are plentiful; see for example[22–24]. A most recent paper that gives a comprehensive overview on metaheuristics is[25].

[1]Department of Biostatistics, University of California, Los Angeles, CA 90095, USA. [2]Alibaba Group, Alibaba, Hangzhou 310099, China. [3]Department of Environmental Science, Tsinghua University, Beijing 100084, China. [4]The Department of Statistics, National Cheng Kung University, Tainan, Taiwan. ✉email: elviscuihan@g.ucla.edu; wkwong@ucla.edu

The motivation of our work comes from our observation that nature-inspired metaheuristic algorithms s are very under-utilized in research in the statistical and life sciences. The aim of this paper is to demonstrate the usefulness of such algorithms to optimize very different types of optimization problems in the statistical and life sciences. As an example, we consider a recently proposed metaheuristic algorithm called competitive swarm optimizer with mutating agents (CSO-MA) by one of the coauthors[26] and demonstrate its utility to solve different types of optimization problems in bioinformatics psychology, ecology, biostatistics and also in the manufacturing industry.

## Nature-inspired metaheuristic algorithms

Nature-inspired metaheuristic algorithms have emerged as a dominant component in the field of optimization[27,28]. These algorithms have gained significant popularity in solving real, high-dimensional, and complex optimization problems. They have found widespread application in engineering, computer science, and various other disciplines to address challenging optimization problems[22,29]. Despite their versatility, these algorithms are under-utilized in some disciplines. One of their key strengths is the availability of widely accessible free codes for users to implement. In addition, they are fast, assumptions-free, and serve as general-purpose optimization algorithms. While they do not guarantee the discovery of an optimal solution, they often yield optimal or near-optimal solutions in a timely manner. Recent studies have demonstrated the ability of swarm-based algorithms to effectively search for previously elusive optimal designs that require solving a 3-layer optimization problem[30]. In the next subsection, we briefly discuss competitive swarm optimization (CSO) and one of its variants.

*Competitive swarm optimizer*
Competitive swarm optimizer (CSO) swarm-based algorithm proposed by[31] and has proven its effectiveness for solving different types of optimization problems with various dimensions . For example[32], applied CSO to select variables for high-dimensional classification models, and[33] used CSO to study a power system economic dispatch, which is typically a complex nonlinear multivariable strongly coupled optimization problem with equality and inequality constraints.

CSO minimizes a given continuous function $f(\mathbf{x})$ over a user-specified compact space $\mathbf{\Omega}$ by first randomly generating a set of candidate solutions. They take the form of a swarm of $n$ particles at positions $\mathbf{x}_1, \cdots, \mathbf{x}_n$, along with their corresponding random velocities $\mathbf{v}_1, \cdots, \mathbf{v}_n$. For tackling design problems, each particle is a candidate design and upon convergence, the solution is the optimal design.

After the initial swarm is generated, at each iteration we randomly divide the swarm into $\lfloor \frac{n}{2} \rfloor$ pairs and compare their objective function values. At iteration $t$, we identify $\mathbf{x}_i^t$ as the winner and $\mathbf{x}_j^t$ as the loser if $f(\mathbf{x}_i^t) < f(\mathbf{x}_j^t)$. The winner retains the status quo and the loser learns from the winner. The two defining equations for CSO are

$$\mathbf{v}_j^{t+1} = \mathbf{R}_1 \otimes \mathbf{v}_j^t + \mathbf{R}_2 \otimes (\mathbf{x}_i^t - \mathbf{x}_j^t) + \phi \mathbf{R}_3 \otimes (\bar{\mathbf{x}}^t - \mathbf{x}_j^t) \qquad (1)$$

$$\text{and } \mathbf{x}_j^{t+1} = \mathbf{x}_j^t + \mathbf{v}_j^{t+1}, \qquad (2)$$

where $\mathbf{R}_1$, $\mathbf{R}_2$, $\mathbf{R}_3$ are all random vectors whose elements are drawn from $U(0, 1)$. The operation $\otimes$ represents element-wise multiplication and the vector $\bar{\mathbf{x}}^t$ is the swarm center at iteration $t$. The social factor $\phi$ controls the influence of the neighboring particles to the loser and a large value is helpful for enhancing swarm diversity (but possibly impacts convergence rate). This process iterates until a pre-specified stopping criterion or criteria are met.

Simulation results have repeatedly shown that CSO either outperforms or is competitive with several state-of-the-art evolutionary and swarm based algorithms, including several enhanced versions of PSO. This conclusion was arrived at after comparing CSO performance with state-of-the-art EAs using a variety of benchmark functions with dimensions up to 5000 and found that CSO was frequently the fastest and with the best quality results[31,34–36].

*Competitive swarm optimizer with mutated agents*
Zhang et al. (2017)[37] proposed an improvement on CSO and call the enhanced version, competitive swarm optimizer with mutated agents (CSO-MA). After pairing up the swarm in groups of two at each iteration, the variant randomly chooses a loser particle $p$ as an agent, randomly picks a variable indexed as $q$ and then randomly changes the value of $\mathbf{x}_{pq}$ to either $\mathbf{xmax}_q$ or $\mathbf{xmin}_q$, where $\mathbf{xmax}_q$ and $\mathbf{xmin}_q$ represent, respectively, the upper bound and lower bound of the $q$-th variable. If the current optimal value is already close to the global optimum, this change will not hurt since we implement this experiment on a loser particle, which is not leading the movement for the whole swarm; otherwise, this chosen agent restarts a journey from the boundary and has a chance to escape from a local optimum. Figure 1 shows the flowchart of CSO-MA. The mutation step (the box in purple) is a key feature of CSO-MA that differentiates it from the standard CSO. The mutation is intended to increase the diversity of the solutions and prevent premature convergence to a local optimum by allowing particles to explore more distant regions of the search space, see[37] for details.

Let $n$ be the swarm size and let $D$ be the dimension of the problem. The computational complexity of CSO is $\mathcal{O}(nD)$ and since our modification only adds one coordinate mutation operation to each particle, its computational complexity is the same as that of CSO. The improved performance of CSO-MA over CSO for finding optimal designs for many complex multi-dimensional benchmark functions has been validated[26]. In the next section, we apply CSO-MA to different estimation problems and show it can produce better quality solutions than conventional methods.

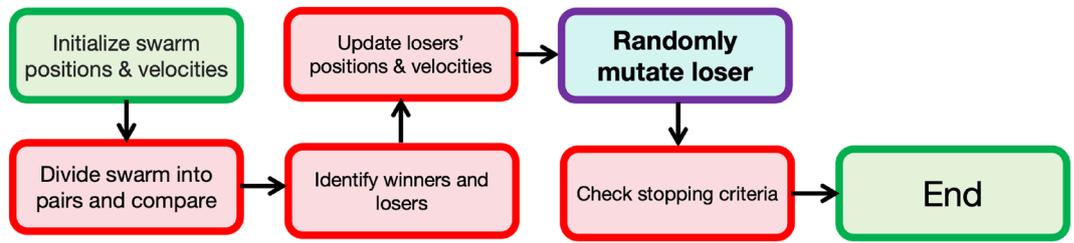

**Figure 1.** Flowchart of CSO-MA.

All computations were performed on a MacBook Pro (16-inch, 2021) with an Apple M1 Max chip and 64GB of memory. The operating system was macOS Sonoma version 14.1.1 and the programming languages were Matlab 2023a and Python 3.9.13. Throughout, the hyper-parameter of CSO-MA was set to $\phi = 0.3$ and the rest of its parameters and those for CSO and PSO are all set to default values. The codes for all the computations are available from the first author by request.

In response to a referee's comment, we show further that CSO-MA is competitive with recently proposed metaheuristic algorithms. As noted in[38], there is a continuing plethora of new or slightly modified proposed as nature-inspired metaheuristics and it is desirable to limit the number of them, unless they are competitive. To this end, we further compare performance of CSO-MA with PSO and CSO using 3 more randomly selected CEC static benchmark functions not used for comparison in[26] and described in[39]. These additional three functions have different mathematical and geometric properties: function $f_9$ is the Weierstrass function (separable), $f_{10}$ is the Quartic function and $f_{11}$ is the Ackley function (non-separable). All three functions have a global minimum of 0, and their optimum were attained at **0** for $f_9$ and $f_{11}$, and $f_{10}$ achieved its optimum at **1**. We also tested the 3 algorithms for their ability to optimize a sphere function $f_{12}$, which is a 2022 CEC dynamic benchmark function. This function was selected at random from the list and is much harder to optimize because it came from a dynamic optimization problem[40,41]). The dimensions of the four functions selected for additional comparison were $D = 100$ and $D = 500$.

Table 1 displays the comparison results after 30 repeated runs. We observe from the table that CSO-MA found the smallest mean values of the optimum when compared with CSO or PSO for $f_9, f_{11}, f_{12}$, but not for $f_{10}$. To test whether there is a significant difference in the medians of the optimal values found by CSO and PSO compared with that from CSO-MA, we applied a Wilcoxon's non-parametric test. Table 2 reports their p-values and suggest that CSO-MA tends to perform more similarly with CSO than PSO in low dimensional optimization problems and that CSO-MA outperforms PSO significantly for the two dimensions tested.

|  | D = 100 (5 for $f_{12}$) | | | | D = 500 (10 for $f_{12}$) | | | |
|---|---|---|---|---|---|---|---|---|
|  | $f_9$ | $f_{10}$ | $f_{11}$ | $f_{12}$ | $f_9$ | $f_{10}$ | $f_{11}$ | $f_{12}$ |
| CSO-MA | | | | | | | | |
| Mean | **1.21E+01** | 8.08E−01 | **1.48E−09** | 4.31E+02 | 7.88E+02 | 4.44E+01 | 1.56E+01 | **5.56E+02** |
| SD | 0.97E+01 | 1.09E−01 | 4.01E−09 | 4.09E+01 | 4.04E+01 | 2.67E+01 | **5.31E−02** | 4.52E+01 |
| CSO | | | | | | | | |
| Mean | 1.81E+01 | 8.76E−01 | 4.98E−08 | 4.39E+02 | 8.04E+02 | 5.08E+02 | 2.64E+01 | 6.00E+02 |
| SD | 1.91E+01 | 4.21E−01 | 1.05E−07 | 5.03E+01 | **2.27E+01** | 5.07E+01 | 6.43E−02 | 7.42E+01 |
| PSO | | | | | | | | |
| Mean | 5.62E+02 | **7.53E−02** | 2.45E−02 | 5.06E+02 | 7.97E+02 | **7.23E+01** | 1.67E+01 | 5.81E+02 |
| SD | 7.55E+01 | 1.23E−01 | 9.19E−02 | 4.55E+01 | 2.89E+01 | **1.05E+01** | 5.05E−01 | **3.39E+01** |

**Table 1.** Performances of the three algorithms for minimizing 3 CEC2008 benchmark static functions ($f_9, f_{10}, f_{11}$) and 1 CEC2020 benchmark dynamic function with multiple optima ($f_{12}$). Significant values are in bold.

|  | D = 100 (5 for $f_{12}$) | | | | D = 500 (10 for $f_{12}$) | | | |
|---|---|---|---|---|---|---|---|---|
|  | $f_9$ | $f_{10}$ | $f_{11}$ | $f_{12}$ | $f_9$ | $f_{10}$ | $f_{11}$ | $f_{12}$ |
| CSO-MA versus CSO | 0.131 | 0.391 | 0.014 | 0.501 | 0.063 | 0.000 | 0.000 | 0.007 |
| CSO-MA versus PSO | 0.000 | 0.000 | 0.000 | 0.000 | 0.325 | 0.000 | 0.000 | 0.018 |

**Table 2.** The *p* values of the Wilcoxon's tests for comparing differences in the medians of the optimized values found from the three algorithms for functions in Table 1. They show CSO-MA outperforms 2 PSO and CSO in 2 of the 3 2008CEC benchmark static functions and the 1 2022CEC benchmark dynamic function algorithms.

## Estimation problems

Metaheuristics has been used to find estimates for model parameters and there is work that showed they can outperform those from statistical packages or find them when the latter fail to do so. For example[42], showed that PSO can find more optimal $L1$-estimates for some models than those in statistical packages. In what is to follow, we demonstrate the CSO-MA can find more optimal maximum likelihood estimates and also able to find them when some statistical packages cannot. Our applications include finding maximum likelihood estimates for models in bioinformatics and research in education, and M-estimates for a Cox regression in a Markov renewal model.

### Single-cell generalized trend model (scGTM)

Cui et al. (2022)[43] proposed a model called scGTM to study relationship between pseudotime[44] and gene expression data. The model assumes that the gene expression has a 'hill' trend along the pseudotime and can be modeled using a set of interpretable parameters. Below is a brief description of the model and shows CSO-MA outperforms PSO algorithm for all but one gene in terms of finding the optimal value of the negative loglikelihood function; details in[43].

For a hill-shaped gene, the scGTM parameters are $\Theta = (\mu_{\mathrm{mag}}, k_1, k_2, t_0, \phi, \alpha, \beta)^T$ and they are estimated from from the observed expression counts $\boldsymbol{y} = (y_1, \ldots, y_C)^T$ and cell pseudotimes $\boldsymbol{t} = (t_1, \ldots, t_C)^T$ using the constrained maximum likelihood method. Here $C$ is the number of cells and the interpretations of the parameters in the model are given in Section 2.1 of[43]. If $\log L(\Theta \mid \boldsymbol{y}, \boldsymbol{t})$ is the log likelihood function, the optimization problem is:

$$\max_{\Theta} \log L(\Theta \mid \boldsymbol{y}, \boldsymbol{t}) \tag{3}$$

such that

$$\min_{c \in \{1,\ldots,C\}} \log(y_c + 1) \leq \mu_{\mathrm{mag}} \leq \max_{c \in \{1,\ldots,C\}} \log(y_c + 1),$$
$$k_1, k_2 \geq 0, \quad \min_{c \in \{1,\ldots,C\}} t_c \leq t_0 \leq \max_{c \in \{1,\ldots,C\}} t_c, \quad \phi \in \mathbb{Z}_+, \tag{4}$$

where

$$\log L(\Theta \mid \boldsymbol{y}, \boldsymbol{t}) = \log \left[ \prod_{c=1}^{C} \mathbf{P}(Y_c = y_c \mid t_c) \right]$$
$$= \sum_{c=1}^{C} \log \left[ (1 - p_c) f(y_c \mid t_c) + p_c \, \mathbb{I}(y_c = 0) \right] \tag{5}$$

and

$$f(y_c \mid t_c) = \frac{\tau_c^{y_c}}{y_c!} \frac{\Gamma(\phi + y_c)}{\Gamma(\phi)(\phi + \tau_c)^{y_c}} \frac{1}{\left(1 + \frac{\tau_c}{\phi}\right)^{\phi}},$$
$$\log(\tau_c + 1) = \begin{cases} b + \mu_{\mathrm{mag}} \exp\left(-k_1(t_c - t_0)^2\right) & \text{if } t_c \leq t_0 \\ b + \mu_{\mathrm{mag}} \exp\left(-k_2(t_c - t_0)^2\right) & \text{if } t_c > t_0 \end{cases},$$
$$\log\left(\frac{p_c}{1 - p_c}\right) = \alpha \log(\tau_c + 1) + \beta,$$

which are all functions of $\Theta$. There are two difficulties in the optimization problem (3). First, the likelihood function (5) is neither convex nor concave. Second, the constraint is linear in $\mu_{\mathrm{mag}}, k_1, k_2,$ and $t_0$ but $\phi$ is a positive integer-valued variable. Hence, conventional optimization algorithms, like P-IRLS in GAM[45,46] and L-BFGS in switchDE[47] are unlikely able to work well. The authors proposed PSO to solve for the constrained MLEs and a Python package is available online. We now apply CSO-MA to the same problem and compare results from the Python package. In addition, we compared CSO-MA's performance with results from two recently proposed metaheuristic algorithms: the prairie dog optimization algorithm (PDO) proposed by[48] and the Rutta and Kutta optimization (RUN) algorithm proposed by[49]. Table 3 displays the negative log likelihood function values found by CSO-MA and PSO for the 20 exemplary genes in[50] after 1000 function evaluations of Eq. (5) for the two algorithms and it shows that CSO-MA outperformed PSO and PDO in all but three of the 20 genes. The Wilcoxon test of CSO-MA against the other two algorithms produced $p$-values less than 0.001 (0.00077 for PSO and 0.00026 for PDO), suggesting that CSO-MA indeed outperformed PSO and PDO in this example.

Figure 2 displays the fitted PAEP gene given by CSO-MA, PSO and PDO. We observe that CSO-MA captures the "fast decreasing trend" when $t \geq 0.8$ better than PSO does, and it reaches the higher peak than PDO does. Figures for other genes also show a consistent pattern.

### Parameter estimation for a Rasch model

The Rasch model is one of the most widely used item response models in education and psychology research[51]. Estimating the parameters in the Rasch and other item response models can be challenging and there is continuing interest to estimate them using different methods and studying the various computational issues. For example[52,53], reported that there are at least 27 R packages indexed with the word "Rasch" and 11 packages capable of estimating parameters and analysis for the Rasch model.

| Gene | CSO-MA | PSO | PDO | Gene | CSO-MA | PSO | PDO |
|---|---|---|---|---|---|---|---|
| PLAU | **1.1115** | 1.1291 | 1.1177 | MMP7 | **1.6562** | 1.6573 | 1.6963 |
| THBS1 | **1.7491** | 1.7498 | 1.7569 | CADM1 | **0.9903** | 0.9907 | 1.0316 |
| NPAS3 | **0.4519** | 0.4598 | 0.4885 | ATP1A1 | **1.0570** | 1.0571 | 1.1407 |
| ANK3 | **1.0473** | 1.0501 | 1.1171 | ALPL | **0.6232** | 0.6235 | 0.6315 |
| TRAK1 | 0.7759 | **0.7758** | 0.7785 | SCGB1D2 | 2.0608 | **2.0501** | 2.0952 |
| MT1F | **0.7851** | 0.7907 | 0.8637 | MT1X | **0.8060** | 0.8065 | 0.9026 |
| MT1E | **0.6580** | 0.6597 | 0.6735 | MT1G | **1.1025** | 1.1414 | 1.1290 |
| CXCL14 | 0.7939 | 0.8512 | **0.7244** | MAOA | **0.8094** | 0.8820 | 0.8161 |
| DPP4 | **0.5503** | 0.5535 | .5528 | NUPR1 | **0.7307** | 0.7854 | 0.7739 |
| GPX3 | **1.7413** | 1.7881 | 1.7904 | PAEP | **2.1034** | 2.3693 | 2.2036 |

**Table 3.** Optimized negative log likelihood (NLL) values (multiplied by $10^5$) obtained by CSO-MA, PSO and PDO after 1000 function evaluations. Lowest NLL values among the three algorithms are in bold for each gene and overall results suggest that CSO-MA outperforms PSO and PDO in almost all cases.. Significant values are in bold.

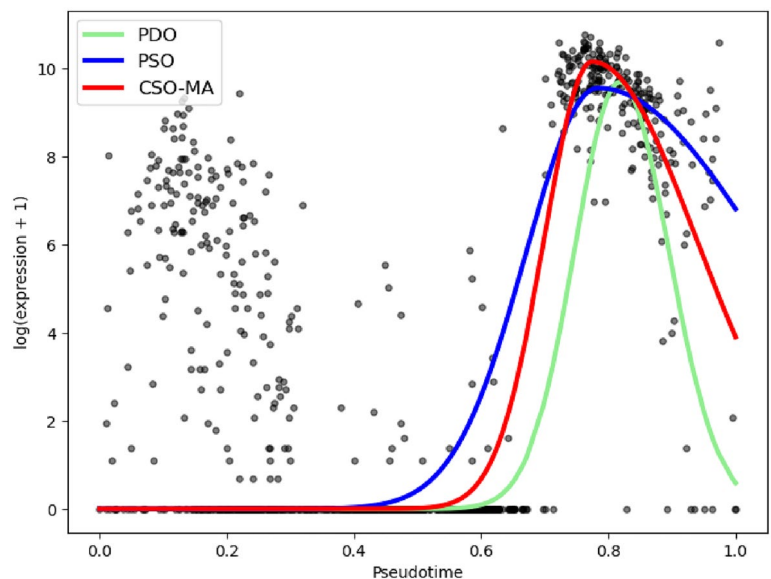

**Figure 2.** Comparison of CSO-MA, PDO and PSO results for the fitted scGTM with gene PAEP.

The expectation-maximization (EM) algorithms is a common method for parameter estimation in statistics[54–56]. The Bock-Aitkin algorithm is a variant of the EM algorithm and is one of the most popular algorithms for estimating parameters in the Rasch models[57]. Because the Rasch model also has many extensions with applications in agriculture, health care studies and in research in marketing[58–60], this subsection compares, for the first time, how metaheuristic algorithms perform relative to the Bock-Aitkin's method.

We give a brief review of the Rasch model before we compare the estimation results given by CSO-MA, Bock-Aitkin's (in the R package *ltm*) and two other metaheuristic algorithms CA and PSO in terms of the likelihood values. In a Rasch model, we work with $N \times I$ binary item response data where 1 indicates correct and 0 indicates incorrect responses. The data come from a cognitive assessment (e.g., math or reading) that includes $I$ test items. A group of $N$ students gave their responses to the $I$ items, and their binary answers to each of the $N$ items were scored and analyzed[51]. The Rasch model is given by:

$$\text{logit}\left(\mathbf{P}(Y_{ji} = 1|\theta_j)\right) = \theta_j - \beta_i, \quad \theta_j \sim N(0, \sigma^2). \tag{6}$$

The item parameter $\beta_i$ represents the difficulty of item $i$ and parameter $\theta_j$ represents the ability of person $j$. We assume that $\theta_j \sim N(0, \sigma^2)$. This model is called the one-parameter model because it considers one type of item characteristic (difficulty). Let $p_{ji} = \mathbf{P}(Y_{ji} = 1|\theta_j)$ and write the marginal likelihood function for model (6) as

$$L(\Theta) = \prod_{j=1}^{N} \int \prod_{i=1}^{I} p_{ji}^{Y_{ji}} (1-p_{ji})^{1-Y_{ji}} \pi(\theta) d\theta, \qquad (7)$$

where $\Theta = (\beta_1, \cdots, \beta_I, \sigma^2)^T$ and $\pi(\theta)$ is the prior of $\theta$.

Metaheuristics has been shown that it can provide superior performance over statistical methods. For instance[61], tackled the challenge of deriving the maximum likelihood estimates for parameters in a mixture of two Weibull distributions with complete and multiple censored data. Their simulation outcomes indicated that the Particle Swarm Optimization (PSO) frequently outperformed the quasi-Newton method and the EM algorithm in terms of bias and root mean square errors.

In this study, we present similar results and show that the nature-inspired metaheuristic algorithm Mutation Algorithm (CSO-MA) can also give more precise maximum likelihood estimates compared to three of its competitors: PSO, the Bock-Aitkin's method, and the Cat Swarm Algorithm (CA). PSO is legendary and an exemplary nature-inspired swarm based algorithm and CA was introduced by[62], and its effectiveness as an optimizer for a single objective function was demonstrated in[63], where they showed its superior competitive edge against several contemporary top-performing algorithms.

We employed the "Verbal Aggression" data set the R Archive[64] and let NLL denote the minimized value of the negative log-likelihood function. Table 4 displays the NLLs from the 4 algorithms, where a swarm size of 30 was used for the 3 metaheuristic algorithms. The hyper-parameter for CSO-MA, was set to $\phi = 0.3$, and the hyper-parameters for PSO and CA were set to the default values in the R package *metaheuristicOpt*[65]. Evidently, CSO-MA has the smallest NLL value and is the winner. The estimated NNL values from CSO-MA, PSO, and Bock-Aitkin are similar, but that from CA is not, suggesting that CA appears less reliable since its estimated NLLs (gold points and lines on the left panel do not come close to the others.

Figure 3 presents a two-panel visualization. The upper panel illustrates the estimated parameters derived from the four algorithms: CSO-MA, Bock-Aitkin, PSO, and CA. Here, the x-axis represents all 24 parameters (encompassing 23 items in addition to the variance parameter) in the model, while the y-axis depicts their estimated values. The lower panel delineates the progression trajectories of the negative log-likelihood functions of the four algorithms, spanning about 100 function evaluations. The left panel shows that except for the CA algorithm, Bock-Aitkin, PSO and CSO-MA give similar parameter estimates; the right panel shows that Bock-Aitkin converges fastest in terms of number of function evaluations while PSO is the slowest. However, CSO-MA has the smallest negative log-likelihood value, or equivalently, the largest log-likelihood value.

### M-estimation for Cox regression in a Markov renewal model

In this subsection, we show CSO-MA can solve estimating equations and produce M-estimates for model parameters, that are sometimes more efficient than those from statistical packages. Askin et al. (2017)[66] correctly noted that metaheuristics is rarely used to solve estimating equations in the statistical community.

| Algorithm | CSO-MA | Bock-Aitkin | PSO | CA |
|---|---|---|---|---|
| NLL | 4038.77 | 4072.93 | 4041.23 | 4780.49 |

**Table 4.** Negative log likelihood values from the four algorithms with CSO-MA outperforming the other three algorithms..

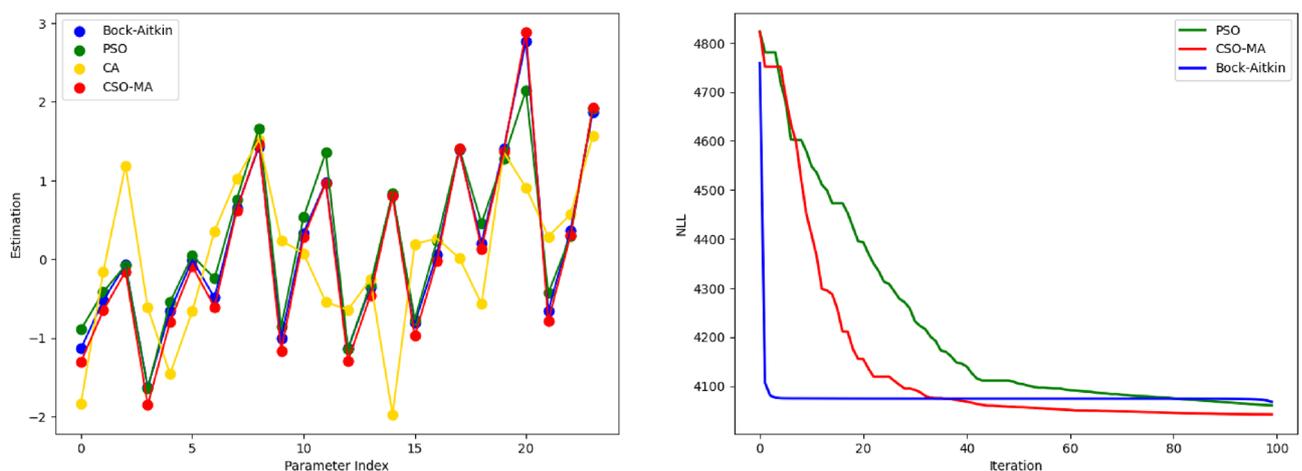

**Figure 3.** The left panel shows estimated parameters from the four algorithms: CSO-MA, Bock-Aitkin, PSO and CA. The x-axis refers to all 24 parameters (23 items plus the variance parameter) in the model and the y-axis refers to the estimated parameter values. The right panel shows the trajectories of the negative log likelihood functions of the four algorithms as they evaluate the negative log-likelihood functions about 100 times..

In a survival study, the experience of a patient may be modelled as a process with finite states[67] and modelling is based on transition probabilities among different states. We take bone marrow transplantation (BMT) as an example. BMT is a primary treatment for leukemia but has major complications, notably Graft-Versus-Host Disease (GVHD), where transplanted marrow's immune cells react against the recipient's cells in two forms: Acute (AGVHD) and Chronic (CGVHD). The main treatment failure is death in remission, often seen in patients with AGVHD or both GVHD types, occurring unpredictably before relapse. The term "death in remission" in the context of leukemia refers to the death of a patient who is in remission from leukemia. This means the patient has achieved remission, where there are no detectable leukemia cells in the body, but they died from other causes that are not directly related to the active progression of leukemia. However, both AGVHD and CGVHD reduce leukemia relapse risks. Hence, there's a five-state model: transplant (TX), AGVHD, and CGVHD are temporary states, while relapse and death in remission are absorbing states[68]. Figure 4 shows the possible transitions among different states (i.e., TX, AGVHD, CGVHD, Relapse and Death).

To model such a process in a mathematically rigorous way, we assume observations on each individual form a Markov renewal process with a finite state, say $\{1, 2, \cdots, r\}$[69]. That is, we observe a process $(X, T) = \{(X_n, T_n) : n \geq 0\}$ where (for simplicity, we do not consider censoring in this subsection), and $0 = T_0 < T_1 < T_2 < \cdots$ are calendar times of entrances into the states $X_0, X_1, \cdots, X_n \in \{1, 2, \cdots, r\}$. In the BMT example, $r = 5$ and $X_n$ takes values in {TX, AGVHD, CGVHD, Relapse, Death in Remission} and $W_i = T_n - T_{n-1}$ represents the sojourn time staying in the state $X_n$. We also observe a covariate matrix $\mathbf{Z} = \{\mathbf{Z}_{ij} : i, j = 1, 2, \cdots, r\}$ where each $\mathbf{Z}_{ij}$ itself is a vector. In practice, we assume that the sojourn time $W_n$ given $X_{n-1} = i$ and $\mathbf{Z}$ has survival probability[70]

$$\mathbf{P}(W_n > x | X_{n-1} = i, \mathbf{Z}) = \exp\left(-\sum_{k=1, k \neq i}^{r} A_{0,ik}(x) e^{\beta^T Z_{ik}}\right)$$

and the transition probability is ($i \neq j$)

$$\mathbf{P}(X_n = j | X_{n-1} = i, W_n) = \frac{\alpha_{0,ij}(W_n) e^{\beta^T Z_{ij}}}{\sum_{k \neq i} \alpha_{0,ik}(W_n) e^{\beta^T Z_{ik}}},$$

where $\beta$ is the parameter of interest, $A_{0,ik}(x) = \int_0^x \alpha_{0,ik}(s) ds$ is the baseline cumulative hazard from state $i$ to state $k$ and $\alpha_{0,ik}(x)$ is the hazard function from state $i$ to state $k$[71]. Suppose we observe $M$ iid individuals and suppose the risk process for an individual is given by $Y_i(x) = \sum_{n \geq 1} \mathbb{I}(W_n \geq x, X_{n-1} = i)$. For a fixed $x$, $Y_i(x)$ counts the number of visits to state $i$ with sojourn time more than $x$ for a particular individual. In the five-state model in Figure 4, since we cannot revisit the states that we have already exited, $Y_i(x)$ is a binary variable. Then from[68,72,73], the estimating equation for $\beta$ is

$$\mathbf{U}(\beta) = \sum_{m=1}^{M} \sum_{i \neq j}^{r} \int_0^\infty \left[ \mathbf{Z}_{ijm} - \frac{S_{ij}^{(1)}(x, \beta)}{S_{ij}^{(0)}(x, \beta)} \right] dN_{ijm}(x). \qquad (8)$$

Here $N_{ijm}(x) = \sum_{n \geq 1} \mathbb{I}(T_n \leq x, X_n = j, X_{n-1} = i)$, $S_{ij}^{(0)}(x, \beta) = \frac{1}{M} \sum_{m=1}^{M} Y_{im}(x) e^{\beta^T Z_{ijm}}$ and $S_{ij}^{(1)}(x, \beta)$ is the first partial derivative of $S_{ij}^{(0)}$ with respect to $\beta$. The M-estimates of $\beta$ are obtained by solving $\mathbf{U}(\beta) = 0$. To apply CSO-MA to obtain the estimates, we turn the problem of solving $\mathbf{U}(\beta) = 0$ into a minimization problem as follows:

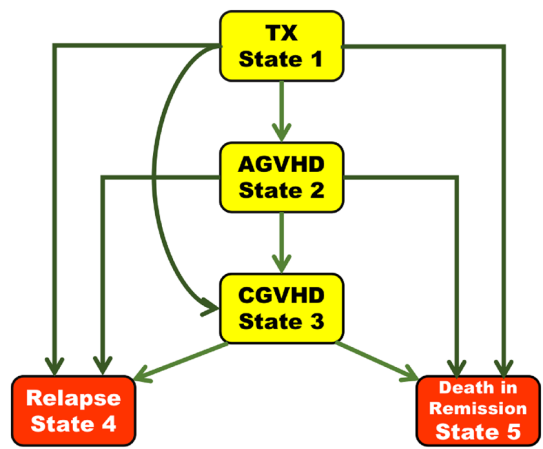

**Figure 4.** A five-state Markov renewal model for BMT failure. Reproduced from[68]. TX = Transplant, AGVHD = Acute Graft-Versus-Host Disease, CGVHD = Chronic Graft-Versus-Host Disease, Relapse = Relapse of leukemia, Death in Remission = Death of a patient who is in remission from leukemia.

$$\widehat{\beta} = \arg\min_{\beta} \|\mathbf{U}(\beta)\|_p \quad (9)$$

where $p \in [1, \infty]$ is a user-selected constant. If the solution exists for $\mathbf{U}(\beta) = 0$, then we have $\min \|\mathbf{U}(\beta)\|_p = 0$ for any $p \geq 1$. Using metaheuristics to creatively solve the system of nonlinear equations[74,75], results from our simulation study suggest that the choice of $p$ does not affect the convergence speed of CSO-MA nor the estimated parameters.

For simulation, we set $p = 2$ and assume $r = 3$, $A_{0,ij}(x) = 0.5x$ for all $i \neq j$, the true parameter vector $\beta = (0.901, 0.759, 0.348)^T$ and elements of the covariance matrix $\mathbf{Z}$ are random uniform variates from $[-1, 1]$. In total, we generated $M = 100$ individuals and the left panel of Figure 5 shows one of the realizations. The swarm size for CSO-MA was 20 and we ran it for 100 function evaluations and the right panel of Figure 5 shows the convergence of CSO-MA. The estimated parameter is $\widehat{\beta} = (0.908, 0.753, 0.329)^T$, which is close to the true value. The observed vector of biases $(0.007, 0.006, 0.017)^T$ is likely due to both the optimization algorithm and the method of partial likelihood itself. The first issue can be reduced by trying using different initialized values of CSO-MA and the second issue may be solved by having a larger sample size so that consistency of the estimators is guaranteed theoretically. For space consideration, we omit additional simulation results that support the effectiveness of CSO-MA for estimating the true parameters correctly.

To further investigate the scalability of CSO-MA and compare it with other algorithms, we perform another simulation study where the state space of $X_i$ consists of two, i.e., $\{1, 2\}$ and 2 is an absorbing state. Consequently, the Markov renewal model is equivalent to a two-state Markov model or a Cox proportional hazards model[71], the sample size is $M = 10{,}000$ and the $\beta$ parameter has is the $100 \times 1$ vector with all entries equal to 1. The elements of the covariance matrix $\mathbf{Z}$ are again generated uniformly from $[-1, 1]$ to mimic the high-dimensional scenario in statistical applications[76]. The simulation is performed on the Matlab 2023a platform. Instead of minimizing the norm of $\mathbf{U}(\beta)$, we minimize the negative partial log-likelihood (NLL) value[68]. We compare CSO-MA with PDO and Runge Kutta optimization (abbreviated as RUN, it is another recently proposed metaheuristics[49]) in terms of their optimum values, stability and running time. The results are given in Table 5. We run each algorithm 30 times to get reasonable statistical results and the number of function evaluation is set to 1000, the swarm size for each algorithm is set to 30. The results suggest that RUN performs the best in terms of NLL and its stability; The CSO-MA has the best performance in terms of average elapsed time and PDO is the slowest among the three algorithms.

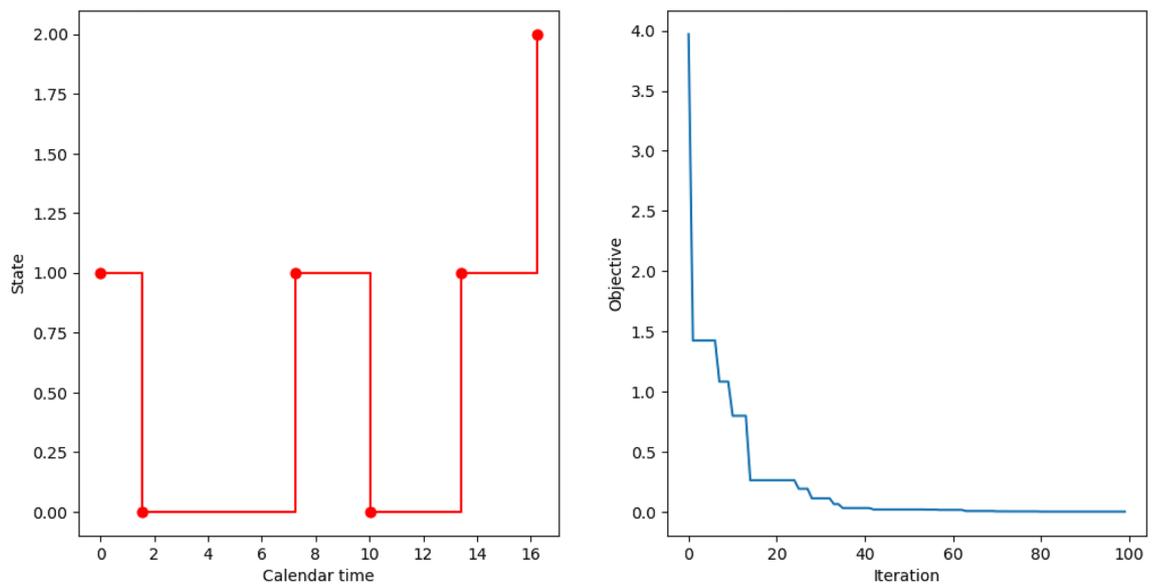

**Figure 5.** Application of CSO-MA to find $M$-estimates for a Cox regression in a Markov renewal model. The left panel is one of the realizations of 100 individuals; the red dots represent the jump times and the transitions for the pair $(X_n, T_n)$. The right panel shows the convergence trajectory of CSO-MA.

| Algorithm | CSO-MA | PDO | RUN |
|---|---|---|---|
| NLL | 1868.51 (248.61) | 1825.20 (2.79) | 1636.39 (4.34) |
| Elapsed time | 250.10s (10.16s) | 472.09s (5.59s) | 270.40s (2.41s) |

**Table 5.** Negative log likelihood values from the three algorithms with CSO-MA outperforming the other two recently proposed algorithms.

**Matrix completion (missing data imputation) in a two-compartment model**
In real studies, such as clinical trials, missing or incomplete data is omnipresent. They occur in computer vision, clinical trials and genomics, just to name a few[77]. Missing data also appear a lot in a recommendation or recommender system, which is defined as a decision making strategy for users under complex information environments[78]; see[79] for an overview of this emerging area of research to alleviate the problem on information load. The best strategy in dealing with missing data is to avoid having them in the first place. This would require constant monitoring of the data and filling in the missing data as soon as they are discovered. Despite the best efforts, missing data abounds and pose problems in data analysis. Matrix completion is the task of filling in the missing entries of a partially observed matrix that represents the data structure. In many instances, the task is equivalent to performing data imputation in statistics. The leads to matrix completion problems and they occur across disciplines. Ensembled models have also been built based on matrix completion for computational drug repurposing to fight the virus SARS-COV-2[80].

In this subsection, we apply CSO-MA to a missing data imputation problem in a non-linear Gaussian regression model using simulated data. Missing data is ubiquitous in all research fields. Imputation is one of the most common ways to fill in and analyze missing data[81] and the Expected Maximization (EM) method[54] is a popular choice for imputing multivariate normal data. We briefly describe the problem and the EM algorithm below.

Suppose that $(Y_1, Y_2) \in \mathbb{R}^2$ has a bivariate normal distribution with mean $\mu(\theta) = (\mu_1(x, \theta), \mu_2(x, \theta))^T$ and a known covariance matrix $\Sigma = \begin{pmatrix} \sigma_1^2 & \rho\sigma_1\sigma_2 \\ \rho\sigma_1\sigma_2 & \sigma_2^2 \end{pmatrix}$ where $\theta$ is a vector of parameters characterizing $\mu$ and $x$ is (possibly) a vector of covariates. We observe $n$ realizations $y_i = (y_{i1}, y_{i2})^T, i = 1, 2, \cdots, n$ and $y_{ij}$ contains missing values for some $i$ and $j$. Let $Y_{(0)}$ and $Y_{(1)}$ denote the observed and missing parts, respectively. On page 250-251 in Little and Rubin (2019)[81], at the $(t + 1)^{th}$ iteration, the $E$ step of the algorithm calculates

$$\mathbf{E}\left(\sum_{i=1}^{n} y_{ij} \Big| Y_{(0)}, \theta^{(t)}\right) = \sum_{i=1}^{n} y_{ij}^{(t+1)}$$

and

$$\mathbf{E}\left(\sum_{i=1}^{n} y_{ij} y_{ik} \Big| Y_{(0)}, \theta^{(t)}\right) = \sum_{i=1}^{n} \left(y_{ij}^{(t+1)} y_{ik}^{(t+1)} + c_{jki}^{(t+1)}\right)$$

for $j, k = 1, 2, \cdots, K$ where

$$y_{ij}^{(t+1)} = \begin{cases} y_{ij} & \text{if } y_{ij} \in Y_{(0)} \\ \mathbf{E}\left(y_{ij} \Big| Y_{(0)}, \theta^{(t)}\right) & \text{if } y_{ij} \in Y_{(1)} \end{cases}$$

and

$$c_{jki}^{(t+1)} = \begin{cases} 0 & \text{if } y_{ij} \text{ or } y_{ik} \text{ is observed.} \\ Cov\left(y_{ij}, y_{ik} \Big| Y_{(0)}, \theta^{(t)}\right) & \text{if } y_{ij}, y_{ik} \in Y_{(1)}. \end{cases}$$

After the $E$-step, missing values are replaced by the conditional expectation derived above. Next, for the $M$-step, we maximize the following conditional log-likelihood with respect to $\theta$ using CSO-MA:

$$\mathbf{E}\left(l(\theta|Y_{(0)}, Y_{(1)}) \Big| Y_{(0)}, \theta^{(t)}\right) = -\frac{1}{2} \sum_{i=1}^{n} \left(\mathbf{y}_i^{(t+1)} - \mu(x_i, \theta)\right)^T \Sigma^{-1} \left(\mathbf{y}_i^{(t+1)} - \mu(x_i, \theta)\right) + C \quad (10)$$

where $\mathbf{y}_i^{(t+1)} = (y_{i1}^{(t+1)}, y_{i2}^{(t+1)})$ and $C$ is a constant independent of $\theta$. Section 8.6 in Wild and Seber (1989)[82] (page 414) illustrates a two-compartment model with (see also chapter 7 in[83])

$$y_{ij} = \mu_j(x_i, \theta) + \epsilon_{ij}, i = 1, 2, \cdots, n, j = 1, 2,$$

where $x$ refers to time, $(\epsilon_{i1}, \epsilon_{i2})^T$ are independently drawn from $N_2(0, \Sigma)$, and the two means are

$$\mu_1(x, \theta) = \theta_1 e^{-\theta_2 x} + (1 - \theta_1)e^{-\theta_3 x}, \mu_2(x, \theta) = 1 - (\theta_1 + \theta_4)e^{-\theta_2 x} + (\theta_1 + \theta_4 - 1)e^{-\theta_3 x},$$

where

$$\theta_4 = \frac{(\theta_3 - \theta_2)\theta_1(1 - \theta_1)}{(\theta_3 - \theta_2)\theta_1 + \theta_2}.$$

Suppose at some time point $x$, the operator failed to record either $y_{i1}$, $y_{i2}$ or both and we observe $Y_{(0)}$ and $Y_{(1)}$ ($n$ observations in total). To make inference about $\theta$, however, we still want to make use of the partially observed data. In this case, we apply the EM algorithm described above to maximize the conditional likelihood (10).

We analyze a real data set to illustrate this idea. The data set comes from Beauchamp and Cornell (1966)[84], see also section 11.2 in Wild and Seber (1989)[82]. We randomly mask some of the values of the data in to be missing in Table 6 and denote them by NA.

Using the complete observations, we estimated the covariance $\Sigma$ to be $\begin{pmatrix} 0.075 & -0.06 \\ -0.06 & 0.06 \end{pmatrix}$ and in the original paper, using full data, the authors' estimated the parameters to be $\hat{\theta} = (0.060, \ 0.007, \ 0.093)^T$. For the EM algorithm, we set the initial $\theta$ to be $(0.381, \ 0.021, \ 0.197)$ and ran CSO-MA for 200 iterations with $\phi = 0.3$. The whole algorithm alternates between computing expression (10) and applying CSO-MA to maximize (10). We ran 10 iterations in total and the imputed results are given in Table 7. We further perform a simulation study (not reported here) with sample size $n = 80$ and 40 missing values in total. The true parameter $\theta$ is $(0.4, 0.05, 0.3)^T$ and the initial value for the EM algorithm is $(0.1, 0.1, 0.1)^T$. The algorithm terminates after 5 iterations, with the estimated parameter value $\hat{\theta} = (0.392, 0.056, 0.275)^T$. This shows that CSO-MA performs well in its role as an optimizer.

### A variable selection problem in ecology

In addition to numerous applications of metaheuristics in engineering and computer science, metaheuristics has also found applications ranging from addressing substantiability issues[85] to land use[19] and agriculture[58]. See also[86], who used metaheuristic algorithms to design placements of the groundwater wells in the Los Angeles Basin.

In this subsection, we apply CSO-MA to a penalized linear regression problem in ecology. Model selection is essential in much of ecology because ecological systems are often too large and slow-moving for our hypotheses to be tested through manipulative experiments at the relevant temporal and spatial scales[87].

The data comes from a plateau lake in Yunnan, China, and was collected by a group of researchers at the Department of Environmental Engineering, Tsinghua University in 2019. They took water samples in March (Spring), June (Summer), September (Autumn) and December (Winter). At each time, 30 sites were sampled from different parts of the waterway. Due to weather issues at the plateau lake in June, data from 6 sites were not recorded. Therefore, the total number of samples is 114 ($= 30 \times 4 - 6$ records). The outcome variable is CRAP and the goal is to determine if and how 17 key variables affect the outcome. Table 8 lists all the regression variables and for space consideration, we only display in the same table, the first two sets of measurements from the $114 \times 18$ data matrix.

Cyanobacteria can form dense and sometimes produce algal toxins. In extreme cases, the cyanobacteria bloom, with high cyanobacterial density or high proportion of cyanobacteria in phytoplankton, can threaten the aquatic ecosystem, fisheries and safety of the water for human drinking. Over the years, the cyanobacterial blooms increase in frequency, magnitude and duration globally[88]. The cyanobacteiral bloom is influenced by the surrounding environment. To effectively control and prevent the cyanobacterial bloom, one of the most important scientific questions is how other factors affect CRAP (Cyanobacteria relative abundance in Phytoplankton).

| i | $x_i$ (h) | $y_{i1}$ | $y_{i2}$ |
|---|---|---|---|
| 1 | 0.33 | NA | 0.03 |
| 2 | 2 | 0.84 | 0.10 |
| 3 | 3 | NA | 0.14 |
| 4 | 5 | 0.64 | NA |
| 5 | 8 | 0.55 | NA |
| 6 | 12 | NA | 0.40 |
| 7 | 24 | 0.27 | 0.54 |
| 8 | 48 | 0.12 | 0.66 |
| 9 | 72 | 0.06 | 0.71 |

**Table 6.** The dataset from Beauchamp and Corenell (1966)[84].

| Imputed data | | | |
|---|---|---|---|
| i | $x_i$ (h) | $y_{i1}$ | $y_{i2}$ |
| 1 | 0.33 | 0.75 | 0.03 |
| 2 | 2 | 0.84 | 0.10 |
| 3 | 3 | 0.65 | 0.14 |
| 4 | 5 | 0.64 | 0.21 |
| 5 | 8 | 0.55 | 0.28 |
| 6 | 12 | 0.39 | 0.40 |
| 7 | 24 | 0.27 | 0.54 |
| 8 | 48 | 0.12 | 0.66 |
| 9 | 72 | 0.06 | 0.71 |

**Table 7.** The imputed dataset.

| CRAP | Depth | Chi-a | DO | Turbidity | pH |
|---|---|---|---|---|---|
| 0.6444 | 0.5 | 34.29 | 6.1 | 4.19 | 9.36 |
| 0.0126 | 0.5 | 18.36 | 6.46 | 15.4 | 9.47 |
| NH4-N | NO3-N | TN | TP | TOC | TDS |
| 0.4 | 0.38 | 0.96 | 0.07 | 22.61 | 906.7 |
| 0.13 | 0.33 | 0.96 | 0.06 | 22.15 | 910.4 |
| T | Ca | K | Mg | Na | F |
| 17.3 | 7.26 | 11 | 68.08 | 191.38 | 2.23 |
| 16.1 | 5.02 | 9.906 | 68.88 | 223.35 | 2.83 |

**Table 8.** Two samples of measurements for the regression variables in the model: Cyanobacteria relative abundance in Phytoplankton (CRAP), the sampling depth of water (Depth), Chlorophyll abundance (Chi-a), dissolved oxygen (DO), turbidity of water (Turbity), potential of hydrogen (pH), Ammonium Nitrogen (NH4-N), Nitrate Nitrogen (NO3-N), total concentration of Nitrogen (TN), total Phosphorus (TP), total organic Carbon (TOC), total dissolve solid (TDS), water temperature (T), Calcium (Ca), Potassium (K), Magnesium (Mg), Sodium (Na) and Fluorine (F).

High values of CRAP often indicate cyanobacterial bloom. Therefore, if we can control the key factors that are associated with CRAP, we can improve environmental health dramatically.

Linear regression analysis is a default choice for detecting association and outliers. We expect that many covariates are correlated. For example, NH4-N and NO3-N are highly correlated with TN. Thus, in reality, some measurements are more important than others to ecologists. In statistics, variable selection and penalized regression methods are proposed to address this issue. In what is to follow, we use CSO-MA and a penalized regression method known as smoothly clipped absolute deviation (SCAD)[89] to selected variables into the model.

Let $y$ be the vector of *CRAP* responses in the linear model, let $X$ be the covariate matrix containing variables *Depth* to $F$ and each column of $X$ is standardize by subtracting its mean and dividing its standard deviation so that each column of $X$ has mean 0 and standard deviation 1. This standardization step is crucial because we want to analyze the relative influence of the variables on *CRAP* and having different scales can cause confusion. All these variables are listed in Table 8. Let $\beta$ be the vector of unknown parameters to be estimated by solving the following optimization problem:

$$\min_{\beta} \|y - X\beta\|_2^2 + \rho \left( \sum_{i=1}^{p} P(\beta_j | \lambda, a) \right), \quad (11)$$

where $\rho$ is the regularization parameter, $a$, $\lambda$ are tuning parameters and

$$P(\beta_j|\lambda, a) = \begin{cases} \lambda|\beta_j| & \text{if } |\beta_j| \leq \lambda \\ \frac{a\lambda|\beta_j| - \beta_j^2 - \lambda^2}{a-1} & \text{if } \lambda < |\beta_j| \leq a\lambda \\ \frac{\lambda^2(a+1)}{2} & \text{if } |\beta_j| > a\lambda \end{cases}$$

is a differentiable and non-convex function and is called the SCAD penalty. The parameter $\rho$ controls the degree of shrinkage applied to the coefficients. A larger $\rho$ increases the penalty on the coefficients, driving them toward zero, and thus, helps in preventing overfitting by enforcing sparsity in the model. We set $a = 2.5$ and $\lambda = 1$, apply SCAD regression to the data ($X$, $y$) for different choices of $\rho$ (see formula (11)) and optimize it using CSO-MA algorithm. We set 12 different values for $\rho$, i.e., $10^{-6}, 10^{-5}, 10^{-4}, 10^{-3}, 0.01, 0.025, 0.05, 0.1, 0.2, 0.5, 1, 10, 100$. For each $\rho$, we record the best particle position found by CSO-MA as our estimation for $\beta$. The CSO-MA algorithm is initialized with 25 particles and iterates 100 times (i.e., 100 function evaluations). We run the algorithm 50 times for each $\rho$ to analyze the stability of CSO-MA. For illustration purpose, we demonstrate the average and standard deviation of the 50 runs when $\rho = 0.025$ and the results are shown in Table 9; further, the average minimum of (11) when $\rho = 0.025$ is 0.315 with a standard deviation of 0.0009 (the other $\rho$'s have similar standard deviation and minimum values), suggesting the stability of CSO-MA algorithm.

Figure 6 illustrates the solution path of SCAD using the CSO-MA algorithm. The x-axis represents the scaled $\rho$ values. When $\rho$ decreases from 100, estimation of *turbidity* (T) deviates from 0 at first. It suggests that turbidity is one of the most important measurements associating with the level of DRAP. One possible reason for such phenomenon is that the turbid water prevents light from penetrating, which in turn indicates a lower amount of the algae carrying out photosynthesis. Further, *temperature* (T) is another variable deviating from 0 at first. The reason is that the optimum temperature for algae growth is $20 + C°$ and the lower the temperature, the less active the metabolism of algaeis. In addition, when $\rho$ decreases from 0.05 to 0.01 ($x$ from 7 to 5), parameter estimation for chemical elements, such as K, Mg, Na, all deviates from 0, suggesting that the concentration of chemical elements has slightly different association of CRAP.

This subsection shows CSO-MA can be usefully applied along with SCAD penalized regression to explore association among different components of water quality and how that affect the outcome CRAP. The interpretation of the solution path is in line with scientific common sense.

| Variable | Average | Standard deviation | Variable | Average | Standard deviation |
|---|---|---|---|---|---|
| Depth | 0.191 | 0.012 | NO3-N | 0.128 | 0.013 |
| Chl-a | − 0.001 | 0.003 | TN | 0.000 | 0.002 |
| DO | 0.219 | 0.015 | TP | 0.047 | 0.008 |
| Turbity | − 0.195 | 0.016 | TOC | − 0.001 | 0.003 |
| pH | − 0.003 | 0.012 | Ca | 0.003 | 0.007 |
| TDS | 0.000 | 0.002 | K | − 0.002 | 0.009 |
| T | 0.499 | 0.033 | Mg | − 0.031 | 0.027 |
| NH4-N | 0.166 | 0.018 | Na | − 0.162 | 0.025 |
| F | − 0.016 | 0.028 | | | |

**Table 9.** Average and standard deviation of parameter estimation after 50 times of runs.

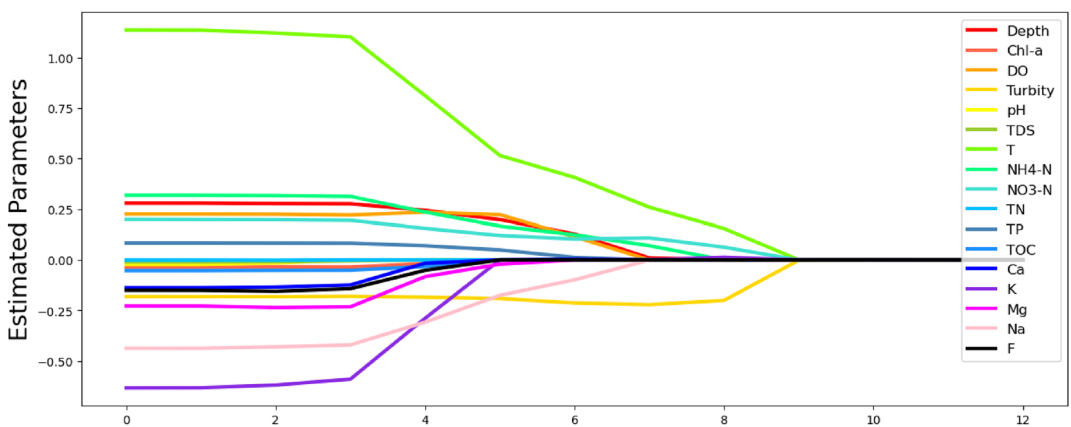

**Figure 6.** Solution path of SCAD using CSO-MA. Each line represents the trajectory of an estimated coefficient for a predictor variable across the ordered values of the regularization parameter $\rho$. The y-axis denotes the estimated coefficient values. The x-axis corresponds to the ordinal position of each $\rho$ value in the set $10^{-6}, 10^{-5}, \ldots, 100$, which have been rescaled to 1, 2, ..., for clarity of presentation.

## Design problems

Design problems are important because experimental costs are always increasing and a well designed study can provide maximum statistical inference precision at minimal cost. Given a design region, a regression model with several factors or independent variables, a design criterion and the total number of observations allowed, an optimal design problem involves finding the optimal number of design points, the optimal combination of factor levels and the proportions of observations to take at the design points. Sometimes the proportions are called weights ($w_i$) and they sum to unity. By working with weights and a convex design criterion, the optimal design problem can be formulated as a convex optimization problem, where theoretical tools are available to confirm whether a solution is optimal. In particular, an equivalence theorem, one for each convex criterion, can be derived using convex analysis results. A by-product is also a design efficiency lower bound that assesses the proximity of a design to the theoretical optimum without knowing the latter. In general, design efficiency is some ratio between 0 and 1 of the criterion value of a design relative to that of the optimum and designs with high efficiencies are sought. If a design has an efficiency of 0.5 or 50%, this means that the design has to be replicated twice to provide the same level of information as the optimal design[90] and[91] provide the technical details.

There are algorithms in the statistics literature for finding optimal experimental designs and even though some of them can be proven to converge mathematically to the optimum[92,93]. However, they may not work well in practice when the model is nonlinear and has several interacting factors. As some of the references below indicated, metaheuristics can outperform traditional algorithms or solve optimization problems that they cannot[94]. For example,[30] solved a standardized maximin optimal design problem and[95] found a minimax optimal design for a random effects hierarchical linear model. Both involved optimizing a non-differentiable objective function and some require multiple nested layers of optimization.

Below is a new application that shows CSO-MA can find a locally $D$-optimal design to estimate all parameters in a logistic model with 10 factors and 3 pairwise interaction terms. $D$-optimal designs are popular because when errors are normally distributed, they minimize the volume of the confidence ellipsoid of the parameters and hence the parameters are accurately estimated. Previous attempts using other metaheuristic algorithms to solve this design problem were less successful because of the large dimension of the optimization problem. For example,[96] applied GA, PSO, CSO to find locally $D$-optimal designs for the Poisson model and logistic model with 5 factors and all pairwise interaction terms[97] used quantum PSO (QPSO) and modified the codes to also find locally *D*-optimal designs. The modified code d-QPSO found a locally *D*-optimal design for a 10-factor logistic model but interactions were not allowed. Likewise,[98] applied differential evolution to find locally *D*-optimal designs for the same model with 5 pairwise interaction terms. However, optimality of their design could not be confirmed but its proximity to the optimum (without knowing the optimum) was assessed using a *D*-efficiency lower bound[90]. The reported design has at least 95% D-efficiency, suggesting they it is close enough to the optimum (without knowing what the optimum is) and likely suffice for most practical purposes.

### Car refueling experiment

[99] described an experiment, based on the logistic model, for testing a vision-based car refueling system with the question that whether a computer-controlled nozzle was able to insert itself into the gas pipe correctly or not[97]. The experiment includes four binary explanatory factors ($x_1 \sim x_4$ numerically taking -1 or 1): ring type (white paper or reflective), lighting (room lighting or 2 flood lights and room lights), sharpening (without or with), smoothing (without or with); six continuous factors ($x_5 \sim x_{10}$): lightning angle (50 to 90 degrees), gas-cap angle 1 (30 to 55 degrees), gas-cap angle 2 (0 to 10 degrees), can distance (18 to 48 inches), reflective ring thickness (0.125 to 0.425 inches) and threshold step value (5 to 15). Experts' opinions suggest that the model should include five specific interaction terms in the model and they are the pairwise interactions terms between ring type and reflective ring thickness, interactions between lighting and lighting angle, interaction between smoothing and car distance, along with 2 3-order interaction terms. To test CSO-MA's potential for finding a locally *D*-optimal design for a more likely realistic model, we include selected interaction terms, namely three two-factor interaction terms and two three-factor interaction terms. Table 10 lists all the terms in the model.

The full model has 10 factors and 16 parameters[97]. Assumed a set of nominal parameter values and found a locally *D*-optimal design using a swarm-based algorithm called Quantum-Behaved PSO (d-QPSO). On average, the runtime for finding the optimal design for the additive linear part of the model without interaction terms, was 140 seconds. They did not report the locally *D*-optimal design for the full model.

We applied CSO-MA to search for locally *D*-optimal designs for both models with and without interaction terms. Since not all factors are likely to interact, we choose to include, as an example, 3 two-factor interaction terms and 2 three-factor interaction terms. We set $k = 20$, which is the initial guess of the number of design points required of the optimal design. We set the number of particles in the algorithm to be $n = 200$ and the stopping criterion is whether the fitness value change is within the pre-specified tolerance value of $10^{-5}$. We ran the algorithm 10 times independently and on average, CSO-MA took 24 seconds to find the same locally *D*-optimal design for the no-interaction model, which is significantly shorter than that required by the d-QPSO employed in[97]. For the model with the interaction terms, CSO-MA was also able to find a locally *D*-optimal design shown in Table 11 with 17 design points and the corresponding weights are in the last column. It has 17 design points, the criterion value is 7.256 and a direct calculation shows its D-efficiency lower bound is 97%. It is not possible to easily confirm optimality of a design for a multi-factor model because it is difficult to visually appreciate the fine features in a high dimensional plot; see[97,98]. Additional numerical checks similar to that described in[96] support that the design found by CSO-MA has all the required features in the equivalence theorem.

### Conclusions

Nature-inspired metaheuristic algorithms are general-purpose optimization tools and they require virtually no assumption for them to work reasonably well. While they are typically used when all other known optimization methods fail, we note that

| Variable | Notation | Type | Range |
|---|---|---|---|
| Ring type | $x_1$ | Binary | – 1 or – 1 |
| Lightning | $x_2$ | Binary | – 1 or – 1 |
| Sharpening | $x_3$ | Binary | – 1 or – 1 |
| Smoothing | $x_4$ | Binary | – 1 or – 1 |
| Lightning Angle | $x_5$ | Continuous | [50, 90] |
| Gas– cap Angle 1 | $x_6$ | Continuous | [30, 55] |
| Gas-cap Angle 2 | $x_7$ | Continuous | [0, 10] |
| Can Distance | $x_8$ | Continuous | [18, 48] |
| Reflective Ring Thickness | $x_9$ | Continuous | [0.125, 0.425] |
| Threshold Step Value | $x_{10}$ | Continuous | [5, 15] |
| P-Interaction 1 | $x_1 x_9$ | – | – |
| P-Interaction 2 | $x_2 x_5$ | – | – |
| P-Interaction 3 | $x_4 x_8$ | – | – |
| T-Interaction 1 | $x_6 x_7 x_8$ | – | – |
| T-Interaction 2 | $x_3 x_4 x_{10}$ | – | – |

**Table 10.** All the model terms in the car refueling study.

| $x_1$ | $x_2$ | $x_3$ | $x_4$ | $x_5$ | $x_6$ | $x_7$ | $x_8$ | $x_9$ | $x_{10}$ | $w$ |
|---|---|---|---|---|---|---|---|---|---|---|
| 1.000 | −1.000 | −1.000 | −1.000 | 50.000 | 30.000 | 0.026 | 31.494 | 0.125 | 5.000 | 0.062 |
| 1.000 | −1.000 | 1.000 | −1.000 | 90.000 | 30.000 | 0.285 | 18.000 | 0.425 | 5.000 | 0.063 |
| 1.000 | −1.000 | 1.000 | −1.000 | 90.000 | 37.342 | 0.000 | 47.999 | 0.425 | 15.000 | 0.061 |
| 1.000 | −1.000 | 1.000 | 1.000 | 68.511 | 55.000 | 0.209 | 29.239 | 0.425 | 15.000 | 0.062 |
| 1.000 | 1.000 | −1.000 | −1.000 | 90.000 | 30.000 | 0.085 | 28.026 | 0.125 | 15.000 | 0.062 |
| 1.000 | 1.000 | −1.000 | −1.000 | 90.000 | 31.591 | 0.000 | 34.269 | 0.425 | 5.000 | 0.062 |
| 1.000 | 1.000 | 1.000 | −1.000 | 50.000 | 55.000 | 0.000 | 33.014 | 0.125 | 5.000 | 0.062 |
| −1.000 | −1.000 | −1.000 | −1.000 | 50.000 | 36.649 | 0.000 | 48.000 | 0.425 | 15.000 | 0.061 |
| −1.000 | −1.000 | −1.000 | −1.000 | 90.000 | 55.000 | 0.025 | 48.000 | 0.425 | 5.000 | 0.062 |
| −1.000 | −1.000 | −1.000 | −1.000 | 90.000 | 55.000 | 0.091 | 36.073 | 0.125 | 15.000 | 0.061 |
| −1.000 | −1.000 | −1.000 | 1.000 | 75.860 | 30.000 | 0.363 | 18.000 | 0.125 | 15.000 | 0.063 |
| −1.000 | −1.000 | 1.000 | −1.000 | 50.000 | 55.000 | 0.007 | 36.516 | 0.125 | 15.000 | 0.062 |
| −1.000 | −1.000 | 1.000 | −1.000 | 90.000 | 30.000 | 0.029 | 38.137 | 0.425 | 15.000 | 0.020 |
| −1.000 | −1.000 | 1.000 | −1.000 | 90.000 | 30.000 | 0.000 | 45.986 | 0.125 | 5.000 | 0.060 |
| −1.000 | 1.000 | −1.000 | −1.000 | 50.000 | 30.000 | 0.000 | 34.471 | 0.125 | 15.000 | 0.057 |
| −1.000 | 1.000 | −1.000 | 1.000 | 67.477 | 30.000 | 0.070 | 48.000 | 0.125 | 15.000 | 0.063 |
| −1.000 | 1.000 | 1.000 | −1.000 | 50.000 | 30.000 | 0.011 | 18.361 | 0.425 | 15.000 | 0.056 |

**Table 11.** A CSO-MA generated locally $D$-optimal design with 17 design points for the 10-factor model with selected interaction terms in the car refueling experiment. The nominal set of values for the model parameters is $\boldsymbol{\theta}^T = (3.0, 0.5, 0.75, 1.25, 0.8, 0.5, 0.8, −0.4, −1.00, 2.65, 0.65, 1.1, −0.2, 0.9, −0.36, 1.07)$.

- improved metaheuristics, such as CSO-MA, can outperform earlier metaheuristic algorithms; this was the case for optimizing parameter estimation in the single-cell generalized trend model, where CSO-MA produced significantly more optimal values than those from two recently proposed metaheuristics;
- they can also produce better quality solutions than those obtained from traditional methods or via commercial statistical packages; this was the case in Table 2 where we observe that the negative likelihood function from the deterministic Bock-Aitkin's algorithm has a larger value than that from PSO and CSO-MA, suggesting metaheuristics outperformed the Bock-Aitkin's procedure;
- the ecological application also demonstrated that results from CSO-MA can have smaller deviations than other estimates, suggesting more stable results from the CSO-MA algorithm compared with standard methods like using CRAP or SCAD;
- improved metaheuristics, such as CSO-MA, can solve optimization problems that were deemed problematic before; such is the case for the car design problem where this paper considers several interacting factors more than earlier papers with a handful of factors in interaction terms.

We close by returning to the question posed in the title of the paper. Based on the current work and other optimization work we have done using metaheuristics, our cumulative experience suggests that they are able to explore and exploit complex optimization problems in statistics and arrive at an optimal solution, or close to the optimum[100]. More interestingly, there are increasing examples that show metaheuristics can outperform statistical optimization methods with theoretical convergence properties in terms of speed or quality of the solution. An example is the frequently used Fedorov's type of algorithms commonly used to generate an optimal design by adding one design point at each iteration, and then periodically collapsing nearby points to a single point heuristically. For this reason, we believe that metaheuristics offers exciting and fertile ground for theoretical and applied researchers and can potentially revolutionize the world of optimization.

## Data availability
The datasets for this paper are available from the corresponding author upon request.

### Acknowledgements
Drs. Wong and Zhang were partially supported by a grant from the National Institute of General Medical Sciences of the National Institutes of Health under Award Number R01GM107639. The content is solely the responsibility of the authors and does not necessarily represent the official views of the National Institutes of Health. The research of Wong was also partially supported by a Yushan Fellowship from the Ministry of Education in Taiwan.

### Author contributions
E.H.C. conceived and worked out Table 1 and all the examples in estimation problems, Z.Z developed the codes for CSO-MA and provided the car refueling experiment example, C.J. C. contributed to the variable selection problem in ecology, and W. K. W. supervised the development of the entire paper, contributed to the literature review, organized the paper, and edited the whole paper. All authors reviewed the manuscript.

### Competing interests
The authors declare no competing interests.

### Additional information
**Correspondence** and requests for materials should be addressed to E.H.C. or W.K.W.

**Reprints and permissions information** is available at www.nature.com/reprints.

**Publisher's note**  Springer Nature remains neutral with regard to jurisdictional claims in published maps and institutional affiliations.

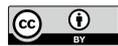 **Open Access**  This article is licensed under a Creative Commons Attribution 4.0 International License, which permits use, sharing, adaptation, distribution and reproduction in any medium or format, as long as you give appropriate credit to the original author(s) and the source, provide a link to the Creative Commons licence, and indicate if changes were made. The images or other third party material in this article are included in the article's Creative Commons licence, unless indicated otherwise in a credit line to the material. If material is not included in the article's Creative Commons licence and your intended use is not permitted by statutory regulation or exceeds the permitted use, you will need to obtain permission directly from the copyright holder. To view a copy of this licence, visit http://creativecommons.org/licenses/by/4.0/.